\documentclass{ieeeaccess}
\usepackage{kotex}
\usepackage{cite}
\usepackage{amsmath,amssymb,amsfonts}
\usepackage{algorithmic}
\usepackage{graphicx}
\usepackage{textcomp}
\usepackage{graphicx}
\usepackage{multirow}
\usepackage{multicol}
\usepackage[table]{xcolor}
\usepackage{hhline}
\usepackage{array}
\usepackage{xcolor}

\def\BibTeX{{\rm B\kern-.05em{\sc i\kern-.025em b}\kern-.08em
    T\kern-.1667em\lower.7ex\hbox{E}\kern-.125emX}}
\begin{document}
\history{Date of publication xxxx 00, 0000, date of current version xxxx 00, 0000.}
\doi{10.1109/ACCESS.2017.DOI}

\title{Challenges in Drone Firmware Analyses of Drone Firmware and Its Solutions}
\author{\uppercase{Yejun Kim}\authorrefmark{1},
\uppercase{Kwnagsoo Cho\authorrefmark{1}, and Seungjoo Kim}.\authorrefmark{1}}
\address[1]{School of Cybersecurity, Korea University, Seoul 02841, South Korea}

\tfootnote{This work was supported as part of Military Crypto Research Center(UD210027XD) funded by Defense Acquisition Program Administration(DAPA) and Agency for Defense Development(ADD) and Supported by a Korea University Grant}

\markboth{IEEE TRANSACTIONS ON MAGNETICS, VOL.~35, NO.~5, SEPTEMBER 1999}
{Author \MakeLowercase{\textit{et al.}}: Preparation of Papers for IEEE TRANSACTIONS and JOURNALS}

\corresp{Corresponding author: Seungjoo Kim (e-mail: skim71@korea.ac.kr).}

\begin{abstract}
With the advancement of Internet of Things (IoT) technology, its applications span various sectors such as public, industrial, private and military. In particular, the drone sector has gained significant attention for both commercial and military purposes. As a result, there has been a surge in research focused on vulnerability analysis of drones. However, most security research to mitigate threats to IoT devices has focused primarily on networks, firmware and mobile applications. Of these, the use of fuzzing to analyze the security of firmware requires emulation of the firmware. However, when it comes to drone firmware, the industry lacks emulation and automated fuzzing tools. This is largely due to challenges such as limited input interfaces, firmware encryption and signatures. While it may be tempting to assume that existing emulators and automated analyzers for IoT devices can be applied to drones, practical applications have proven otherwise. In this paper, we discuss the challenges of dynamically analyzing drone firmware and propose potential solutions. In addition, we demonstrate the effectiveness of our methodology by applying it to DJI drones, which have the largest market share.

\end{abstract}

\begin{keywords}
Drones, Firmware, Dynamic Analysis, Fuzzing
\end{keywords}

\titlepgskip=-15pt

\maketitle

\section{Introduction}
\label{sec:introduction}
\PARstart{A}{dvances} in drone technology have allowed drones to be utilized in a variety of industries. Because they are relatively small and can be operated unmanned, drones are actively used in industries such as construction sites, agriculture, perimeter surveillance, and military operations that are difficult or dangerous for humans to perform[1], [2]. However, there have been cases in the U.S. in the past when the Department of Defense's drone system was hacked and all information from the drones was stolen, and hackers took control of the drones and interfered with military operations[3]. There have also been cases where malicious actors have used drones to interfere with the takeoffs and landings of nearby aircraft, causing flights to be canceled. As such, drones have a wide range of applications, and vulnerabilities can lead to personal information leakage and disruption of key facilities. In response to these incidents, the U.S. Navy held a vulnerability analysis competition for U.S. Navy drones called "Hack the Sky" to strengthen the security of drones[4]. While there are many activities and studies to enhance the security of drones, there is a lack of research on dynamic analysis such as fuzzing of drone firmware.

As drones are utilized in various fields, various functions are added to the drone firmware and the structure of the firmware is becoming more complex. Manually reading and analyzing the source code of complex and large software by humans can lead to incorrect analysis results as the size and structure of the software to be analyzed increases[5]. Existing publicly available firmware analysis tools are limited to routers, IP cameras, and mobile applications, making them difficult to use for drone analysis. In addition, firmware analysis studies to improve drone security are limited to specific models, such as DJI's Phantom series, making them less useful[6], [7], [8]. 

Also existing researches about drone dynamic analysis are mainly related to communication protocols used in drone systems such as MAVLink and DSMX [34], [35], [36]. However, it is not just attacks on network protocols that are used to attack drones, but also vulnerabilities within the system. Therefore, in addition to the existing research, an analysis of the firmware binary used in drones should be carried out. In the case of drones, there are no simple input interfaces, such as Linux terminal or SSH(Secure Shell). Therefore, in order to perform the drone dynamic analysis with existing firmware dynamic analysis tools, it is necessary to make them recognize the drone input interface for automatic data injection. In addition, attacks using communication protocols require the attacker to be within the communication range of the drone, which is very difficult to do if technologies such as communication encryption and anti-jamming are applied. And even if internal access is possible through attack surfaces such as communication protocols, if no vulnerability analysis is performed on the actual firmware binaries, we are forced to rely on simple mutation-based analysis. 

Embedded systems like drones, which often run on Linux-based firmware, are especially susceptible to a spectrum of cybersecurity threats, ranging from denial of service and unauthorized root-level access to the potential for arbitrary code execution[37]. Attackers have exploited these vulnerabilities to inject malware into devices and build botnets to launch attacks [38]. In addition, external wireless communication opens avenues for attacks through internal firmware libraries and application-internal functions. Research from the SMACCM project within the U.S. Defense Advanced Research Projects Agency's (DARPA) HACMS demonstrates the potential risk of planting malware with physical access to a weapon system, highlighting the need for a mathematically secure operating system [39]. In one notable experiment, seL4, a secure operating system tailored for this purpose, was deployed on a real helicopter. Despite being tested with an attack through a USB port aimed at crashing the helicopter, seL4 withstood all physical access attacks, demonstrating its resilience [40]. Consequently, our study addresses the dynamic analysis of internal firmware binaries in the context of drone security, complementing the current focus on protocol and internal memory forensics.

The DJI, which has a share of more than 70\% of the global drone market, encrypts and releases very many kinds of drone firmware with its own code signature. So, it is difficult for firmware analysis tools to recognize and analyze the instruction codes in the DJI drone’s firmware. To avoid these problems, we propose a method for automatic dynamic analysis of increasingly complex and large drone firmware, and the challenges and solutions that may exist. In doing so, we aim to expand research into dynamic analysis of drone firmware and improve security in this area. In particular, we identified three major challenges:
\begin{itemize}
    \item Firmware acquisition and decryption: How to acquiring the drone firmware and making it analyzable?
    \item Operating environment issues with existing analysis tools: Drone system has poor performance to perform on-board dynamic analysis
    \item Library dependency issues: What are the necessary components for the construction of the dynamic analysis environment for drone firmware?
\end{itemize}
A detailed description of each challenge is provided in 3.A.

This paper consists of four chapters: Chapter 2 describes the research related to firmware analysis, and Chapter 3 discusses the problems and solutions for analyzing drone firmware. Chapter 4 concludes with the limitations of this research and future research directions.

\subsection{OUR APPROACH}
In this paper, we introduce a methodology designed to address the challenges of dynamic analysis for drone firmware, with a focus on DJI drones, which hold a significant market share. Our approach streamlines the dynamic analysis process by automating the complex firmware decryption and creating an emulation environment that accurately mimics the drone's operational conditions. This significant simplification enables more efficient security research on drone firmware, reducing the time and effort traditionally required.

Particularly, we employ AFL (American Fuzzy Lop) for fuzzing techniques to effectively identify potential security vulnerabilities within the binary code of drone firmware. AFL is automated tool to generating and testing of input cases significantly increases test productivity. This automation, combined with its open source nature, allows for continuous improvement and adaptability to different test environments. In addition, the tool's ability to reproduce crashes and log critical test cases facilitates efficient debugging and problem resolution. AFL is s proven track record, combined with its advanced features and community support, make it an ideal choice for fuzz testing in our research[45]. This step is crucial for enhancing the security of drone firmware and is expected to significantly contribute to the safe operation of drones.

Additionally, to facilitate knowledge sharing within the research community, we have made the tools we developed open-source. This allows other researchers to easily apply our methodology to their own studies, lowering the barriers to entry for drone firmware security research and aiming to cultivate a safer drone technology ecosystem. This research provides a robust foundation for innovatively advancing dynamic analysis and security research on drone firmware. In summary, our contributions to meeting the challenges outlined above are as follows:
\begin{itemize}
    \item A step-by-step guide on how to perform dynamic analysis for drones, which includes firmware acquisition, decryption, dynamic analysis, and building analysis tools suitable for the environment.
    \item How to overcome performance and power resource issues when conducting dynamic analysis on systems with limited resources and challenging environments.
    \item Set up an operating system for dynamic analysis of drone firmware, including manual steps such as finding dependent libraries and selecting the appropriate operating system version.
\end{itemize}

\section{Related Works}
In this chapter, we elucidate the methodology for drone analysis through a review of existing research findings related to automatic firmware analysis and vulnerability assessment targeted at drones. Moreover, we explore existing tools for approaches to firmware analysis. Current tools for drone analysis predominantly focus on evaluating vulnerabilities within communication protocols or forensic applications to retrieve internal metadata. While there is existing research on conducting dynamic analysis of drone firmware, its applicability is generally limited to open-source drone platforms like ArduPilot[35]. This utility is constrained when source code is not accessible. Hence, there is a pressing need for the development of dynamic analysis methodologies applicable to firmware binaries. In order to collect and examine previous studies, we has drawn upon established survey-style research methodologies[9]. 

\subsection{Drone Vulnerability Analysis}
Recently, as drones have become important for commercial and military purposes, various vulnerability analysis studies on drones are also being conducted. However, to the best of our knowledge, most of them are analyzing vulnerabilities in communication protocols between drones and controllers, and no research has been conducted to automatically analyze the firmware of drones. The current research on drones looks like this Figure 1.

\Figure[t!](topskip=0pt, botskip=0pt, midskip=0pt)[scale=0.55]{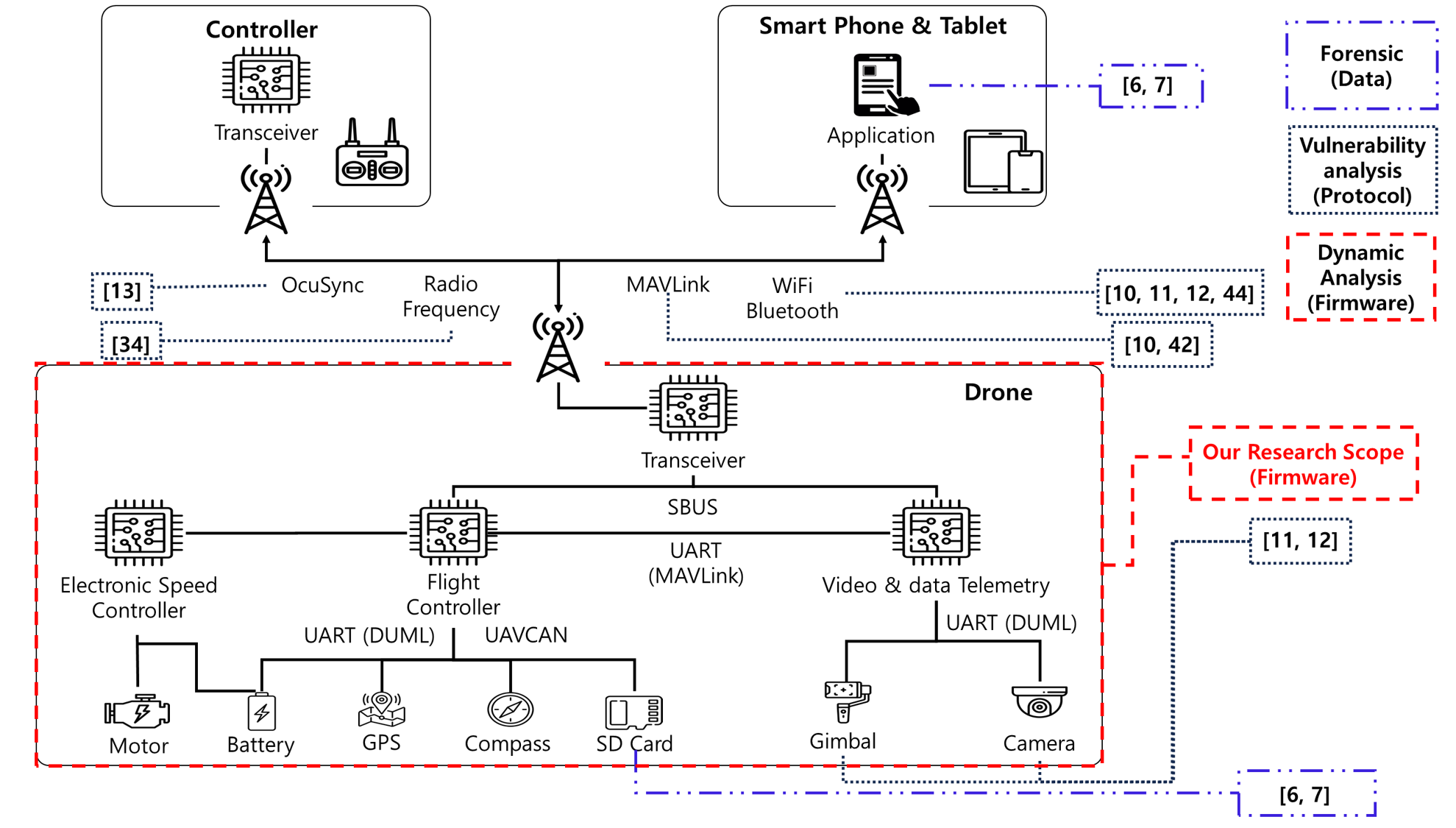}
{Related research on drone analytics\label{fig1}}

For example, Kwon in their 2018 IEEE Access paper "Empirical analysis of MAVlink protocol vulnerability for attacking unmanned aerial vehicles", proposed an attack methodology that exploits the MAVlink protocol vulnerability commonly used in unmanned aerial vehicles (UAVs)[10]. And Fadilah unveiled a drone-targeted vulnerability assessment tool called DRAT at the 2020 ACM Conference on Data and Application Security and Privacy. DRAT works on the wireless signals used by drones, including replay attacks and GPS spoofing[11]. Later, the author added features such as attacking 802.11 protocol targets and providing a GUI to the existing DRAT, and released it through ICIEA in 2021[12]. Schiller designed and implemented a decoder that works with DUML, a proprietary tracking protocol used exclusively by DJI, in "Drone Security and the Mysterious Case of DJI's DroneID," published in NDSS in 2023[13]. At Recon23 in 2023, the same author delivered a presentation titled "Unchained Skies: A Deep Dive into Reverse Engineering and Exploitation of Drones," focusing on reverse engineering and exploiting DJI drones. This talk primarily addressed signal analysis, firmware modifications, and PCB analysis related to DJI drones. The author highlighted that due to the drones' intricate firmware architecture, which does not operate on a singular binary system, emulation is challenging. Additionally, the absence of publicly available source code renders existing fuzzing tools ineffective for such purposes[41]. Recently, Nozomi Networks published the results of a vulnerability analysis against the DJI Mavic 3 drone [44]. However, again, the vulnerabilities were identified based on static analysis of the DJI Mavic 3's mobile application and firmware, and fuzzing was performed by targeting the Wi-Fi protocol. Emulation and fuzzing are also performed here, but since we are dealing with communication protocol-based vulnerabilities, dynamic analysis of the firmware binary is not covered.

As such, the majority of drone analysis tools currently available target vulnerabilities in drone communication protocols and fall short when it comes to firmware analysis. The objectives and approaches employed in prevailing research are outlined in Table 1.

\begin{table}
\caption{Goals and methods of achieving existing drone studies}
\label{table}
\setlength{\tabcolsep}{3pt}
\begin{tabular}{|>{\centering\arraybackslash}p{30pt}|>{\centering\arraybackslash}p{87pt}|>{\centering\arraybackslash}p{114pt}|}
\hline
\textbf{Ref}& 
\textbf{Goal}&
\textbf{Method}\\
\hline
% DROP (DRone Open source Parser) your drone: Forensic analysis of the DJI Phantom III
{[6]}& 
Developing forensic tools

(Smart Phone, SD Card)&
Developed a tool to perform forensics on drones by extracting encrypted/encoded DAT and TXT files and metadata stored on the drone's SD card and mobile device.\\
\hline
% Drone forensics: A case study on DJI phantom 4.
{[7]}&
Forensic method suggestions

(Smart Phone, SD Card)&
Conducted research on forensic methods to extract data about internal flight logs, photos, etc. via the smartphone device's backup utility and SD card.\\
\hline
%Empirical analysis of mavlink protocol vulnerability for attacking unmanned aerial vehicles 
{[10]}&  
Protocol vulnerability analysis

(GCS, MAVLink)&
Conducted research on stopping UAVs by flooding ICMP over the MAVLink protocol to stop them.\\
\hline
%DRAT: A drone attack tool for vulnerability assessment
{[11], [12]}& 
Developing vulnerability analysis tools

(RF, GPS)&
Developed vulnerability analysis tools using Software Defined Radio (SDR) based retransmission attacks, targeted target identification using WiFi, GPS spoofing, and side-channel attacks using USB ports.\\
\hline
%Drone Security and the Mysterious Case of DJI’s DroneID
{[13], [41]}& 
Protocol vulnerability analysis

(Protocol, UART(DUML))&
Conducted research on communication signal analysis of DJI drone, bootloader analysis using UART, and fuzzing method of communication protocol between smartphone and drone based on DJI drone's communication protocol (DUML).\\
\hline
%Security Analysis of Drone Communication Protocols
{[34]}& 
Protocol vulnerability analysis

(DSM)&
Conduct research on how Denial of Service or taking over control attacks can be performed based on DSM protocols.\\
\hline
%Security analysis of the drone communication protocol: Fuzzing the MAVLink protocol
{[42]}& 
Protocol vulnerability analysis

(MAVLink)&
Conducted research on how protocol fuzzing could be performed based on the MAVLink protocol used in two-way communication between the drone and GCS.\\
\hline
 {[44]}& Protocol vulnerability analysis 
 
 (Wi-Fi)&Static analysis of mobile applications and firmware to find, emulate, and fuzz Wi-Fi vulnerabilities\\\hline
 Proposed works& 
 Firmware Dynamic Analysis  
 
 (Firware binary)&
 Using emulation and fuzzing techniques, we are researching how to perform binary dynamic analysis on drone firmware.\\\hline
\end{tabular}
\label{tab1}
\end{table}

\subsection{IoT Firmware Vulnerability Analysis}
Several vulnerability analysis tools that target firmware of IoT devices, but not drones, have been published, but their application to drones is problematic. Research on vulnerability analysis tools for IoT devices is mainly focused on networks, mobile applications, or analysis targets such as routers and IP cameras. In addition, in order to perform dynamic analysis for each product, it is necessary to have a separate operating environment that can emulate each product. However, most IoT devices that run firmware are characterized by a high degree of hardware dependency. Because of this, existing firmware analysis tools are specialized for the products they target, making them unsuitable for dynamic analysis of other products'(such as drone) firmware.

\section{Dynamic Drone Firmware Analysis: Challenges and Solutions}
This chapter describes the entire process of analyzing drone firmware. In this study, we limit the scope of our research to DJI drone systems. DJI currently has the highest share of the drone market, which is over 70\% [21]. We performed a dynamic analysis based on the firmware used in DJI drones, and we would like to elaborate on the challenges we encountered during the analysis and our efforts to solve them. The overall process we followed to dynamically analyze the drone firmware is shown in Figure 2.

\Figure[t!](topskip=0pt, botskip=0pt, midskip=0pt)[scale=0.75]{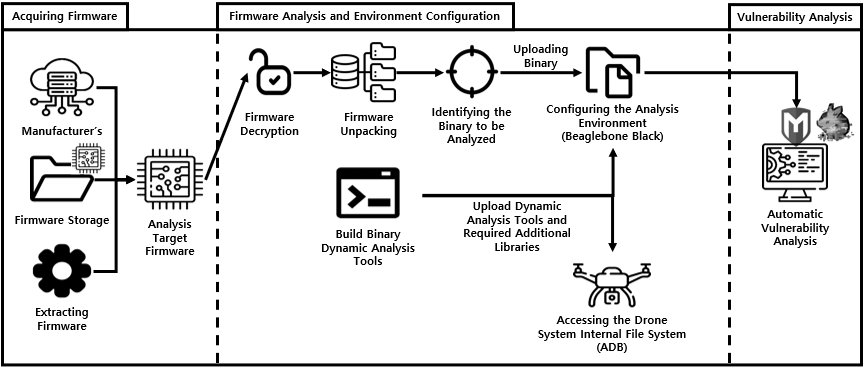}
{Drone firmware dynamic analysis process\label{fig2}}

\subsection{Challenges of drone firmware analysis}
\textbf{Firmware acquisition and decryption.} In order to perform dynamic analysis on drone firmware, we need to collect drone firmware from the DJI company we have selected for analysis. To collect drone firmware, we can either extract it directly from the drone device or collect publicly available data. If you were to access the inside of a DJI drone via ADB and extract the firmware, you could extract the unencrypted binary, but only if you physically have the drone, and you would need to exploit a vulnerability such as a race condition to enable a disabled ADB feature. In addition, it is possible to obtain the drone firmware through existing published research, but the firmware of DJI drones is encrypted with DJI's signature and AES, which must be decrypted[43].

\textbf{Operating environment issues with existing analysis tools.} For the collected firmware, we used emulators designed for existing firmware analysis or installed existing binary dynamic analysis tools such as GDB server and fuzzer directly on the drone system. However, the emulator or analysis tool designed for existing firmware analysis could not be used due to issues such as dependencies in the environment and firmware libraries and differences in the operating environment. In addition, the method of installing dynamic analysis tools directly on the drone system was successful [13], but there is the problem of not having enough time for dynamic analysis due to limited battery capacity. Drones are designed to operate on battery power, which results in a limited operating time. This limitation makes it difficult to perform comprehensive and time-consuming dynamic analysis on specific programs within the given time frame. Even if the drone system has uninterrupted power supply, the constrained hardware environment, including limited memory and processing power, hinders the efficient execution of dynamic analysis tools, which affects the effectiveness of the analysis. At this time, it is also possible to consider using multiple sampling, which can obtain more efficient results by sampling various input data several times. However, tools such as MULTIFUZZ[46], PosFuzz[47], Honggfuzz[48], and libFuzzer[49], which are fuzzers using the multiple sampling technique, were not available due to problems such as operating environment, as a result of using them in this study. Furthermore, multiple sampling cannot be effectively applied in situations where CPU and memory limitations exist. Furthermore, the battery usage speed could be accelerated, which made it challenging to utilise the multiple sampling technique. This technique is more suitable for dynamic analysis in an emulation environment, and it is more suitable to apply to tools in emulators or software than drone hardware.

\textbf{Library dependency issues.} In order for the dynamic analysis tool for analyzing drone firmware to run, the corresponding program must first be executed. It is also important that the libraries referenced by each program are installed in the same location on the file system for the program to run. In our analysis environment, we have verified that the drone firmware architecture works and that some library-less binaries can work. However, there is a problem with finding the correct library that most drone firmware binaries refer to.

We undertook research to address these issues in order to facilitate dynamic analysis of drone firmware. The details of each problem and what we did to address them are discussed in Section 3.2.

\subsection{Our solutions for drone firmware dynamic analysis}
We propose a method for the dynamic analysis of drone firmware and conduct a study of the problems and solutions encountered during the analysis process. Each problem described in Section 3.1 is discussed in the following subsections. Section 3.2.1 describes the process of acquiring the drone firmware and decrypting the encrypted firmware. Section 3.2.2 describes the use of embedded cards to solve the problem of not being able to use existing analysis tools. Section 3.2.3 describes how to extract the necessary libraries from the actual drone to run the drone firmware binary and place them on the embedded card to solve the library dependency problem. Finally, Section 3.2.4 covers the setup process for performing fuzzing within the embedded board built to analyze the drone firmware.

\subsubsection{Acquiring the binary file for analysis}
This section describes how to obtain the drone firmware for analysis. If a physical drone exists, the method of extracting the firmware directly from the drone system can be used to obtain the firmware. Drone systems are primarily implemented based on the Android operating system. The Android operating system supports the Android Debugging Bridge (ADB) for debugging, and if a physical drone exists, the ADB can be used to obtain binary files within the drone system [26].

\Figure[t!](topskip=0pt, botskip=0pt, midskip=0pt)[scale=0.73]{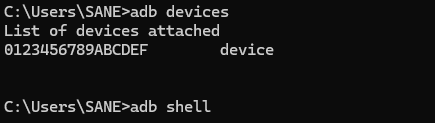}
{Connecting the DJI drone to ADB\label{fig3}}

The advantage of directly extracting the firmware by connecting to the physical drone with ADB is that the internal signature and encryption issues can be resolved, and the extracted firmware can be analyzed immediately. However, analyzing and extracting different firmware versions requires a physical drone of each type, and there are cases where the ADB connection is limited. Therefore, we have collected publicly available DJI drone firmware. However, drone firmware is not officially available for download as a separate firmware file, so we need to extract the firmware directly from the drone or use publicly available drone datasets for our research. We used publicly available DJI drone firmware in a repository called DankDroneDownloader (DDD) and compared it to firmware extracted using ADB to ensure the reliability of the firmware. We were able to obtain 762 drone firmware from DDD, but they were all signed and encrypted.

The DJI drone firmware analyzed in this research was encrypted with AES and RSA, making it impossible for existing analysis tools to read the firmware. To solve this problem, it is necessary to decrypt the firmware file with the correct key. The decryption keys for DJI drone firmware are publicly available from the DJI drone research team "o-gs" (Original Gangsters)[23]. In addition, Tienen, a member of o-gs, conducted research on extracting AES encryption keys through side-channel analysis in 2022 and published decryption keys for some DJI drones [22], [25]. Although decryption keys for DJI drone firmware are thus publicly available, the process of finding the appropriate decryption key for each firmware file from the binary header and inserting the firmware into a decryption tool is a simple iterative process performed through a one-to-one mapping, which can be solved more efficiently by automation than manually. Therefore, in order to automate the firmware decryption module to be applied to the firmware analysis tool to be developed, we conducted research and developed an automatic firmware decryption module in the following order.\\

\begin{enumerate}
    \item Acquiring the drone firmwares.
    \item Analyze the structure of the firmware file.
    \item  Based on the analyzed firmware file structure, identify areas where information about the decryption key can be obtained.
    \item Obtain the decryption key corresponding to the identifier.
    \item Configure a dictionary with the identifier as key and the decryption key as value.
    \item Create and apply a firmware auto decryption module based on the configured dictionary.\\
\end{enumerate}

For our study on automating drone firmware analysis, we collected a total of 764 firmware for 50 DJI products. The firmware we collected were obtained from publicly available online sites such as search engines like Google, Github, and manufacturer sites. The collected firmware is shown in Table 2.\\

\begin{table}
\caption{Collected DJI firmware}
\label{table}
\setlength{\tabcolsep}{3pt}
\begin{tabular}{|>{\centering\arraybackslash}p{190pt}|>{\centering\arraybackslash}p{40pt}|}
\hline
\textbf{DJI Firmware Model}& 
\textbf{Count}\\
\hline
A3 Flight Controller& 
14\\
AC202& 
7\\
Action 2&
2\\
Aeroscope&
2\\
AG411&
1\\
AG600 Gimball&
4\\
AGRAS Series&
41\\
Avata&
2\\
Crystalsky 5.50 Inch&
1\\
Crystalsky 7.85 Inch&
2\\
DJI Battery Station&
1\\
DJI RC &
6\\
DJI RC Pluse&
1\\
DJI RC Pro&
5\\
DJI Transmission&
1\\
D-TRK GNSS&
9\\
FPV Goggles V1&
7\\
FPV Goggles V2 (DIY FPV Mode)&
4\\
DJI Racer&
15\\
FPV System - Air Unit&
7\\
FPV System - Air Unit Lite&
5\\
FPV System - RC&
7\\
Goggles - Series&
24\\
Inspire 1&
66\\
Inspire 1 Pro&
14\\
Inspire 2&
12\\
Matrice 30 Series&
2\\
Matrice 100&
11\\
Matrice 200 Series&
19\\
Matrice 300 Series&
15\\
Matrice 600 Series&
61\\
Matrice M30 M30T&
1\\
Mavic Air Series&
42\\
Mavic Mini Series&
31\\
Mavic Pro 1 Series&
31\\
Mavic Pro 2 Series&
40\\
Mavic Pro 3 Series&
12\\
N3 Flight Controller&
20\\
Ocusync Air System&
10\\
Osmo Series&
20\\
Phantom 2 Series&
2\\
Phantom 3 Series&
94\\
Phantom 4 Series&
44\\
Phantom 4 RTK - China Only Version&
1\\
Robomatser S1&
16\\
Smart Controller&
8\\
Spark&
16\\
WM222&
1\\
Zemnuse Series&
7\\
\hline
Total&
764\\

\hline
\end{tabular}
\label{tab1}
\end{table}

To decrypt the DJI firmware, you need information about the encryption method used in the firmware and the decryption key. To obtain this information, we first need to analyze the structure of the firmware file to identify the areas where hints about the encryption scheme and decryption key are located.

The binary header of the firmware is usually located at the beginning of the file and contains information such as the type, version, and date of the firmware. Most IoT products retrieve information from this binary header to recommend a firmware upgrade to the user or to determine if the firmware is compatible. However, DJI's firmware has its own signature and encryption for enhanced security.

Decryption is performed by reading the header information from a decryption module mounted on the drone's hardware chip[24]. Therefore, we analyzed the relevant research to collect the header information that DJI drones need to run the firmware[25].

According to the literature on DJI firmware, the DJI drone firmware header contains information about the magic number, version, authentication key identifier, scramble key, etc. Inside the drone, the header information of the corresponding firmware binary is used to verify and decrypt the signature value. The entire DJI drone firmware header is organized as follows in Table 3.
\\

\begin{table}
\caption{Header in DJI drone firmwares}
\label{table}
\setlength{\tabcolsep}{3pt}
\begin{tabular}{|>{\centering\arraybackslash}p{57pt}|>{\centering\arraybackslash}p{57pt}|>{\centering\arraybackslash}p{114pt}|}
\hline
\textbf{4-Byte}& 
\textbf{4-Byte}&
\textbf{8-Byte}\\
\hline
Magic ("IM*H")& 
Version&
Unknown
\end{tabular}
\begin{tabular}{|>{\centering\arraybackslash}p{57pt}|>{\centering\arraybackslash}p{57pt}|>{\centering\arraybackslash}p{54pt}|>{\centering\arraybackslash}p{54pt}|}
\hline
\textbf{4-Byte}& 
\textbf{4-Byte}&
\textbf{4-Byte}&
\textbf{4-Byte}\\
\hline
Header size& 
RSA sig size&
Payload size&
Unkown\\
\end{tabular}
\begin{tabular}{|>{\centering\arraybackslash}p{120pt}|>{\centering\arraybackslash}p{54pt}|>{\centering\arraybackslash}p{54pt}|}
\hline
\textbf{8-Byte}& 
\textbf{4-Byte}&
\textbf{4-Byte}\\
\hline
Unknown&
Auth key identifier&
Encryption key identifier\\
\end{tabular}
\begin{tabular}{|>{\centering\arraybackslash}p{240pt}|}
\hline
\textbf{16-Byte}\\
\hline
Scramble key\\
\hline
\textbf{32-Byte}\\
\hline
Image name\\
\hline
\textbf{48-Byte}\\
\hline
Unknown\\
\hline
\end{tabular}
\begin{tabular}{|>{\centering\arraybackslash}p{180pt}|>{\centering\arraybackslash}p{54pt}|}
\textbf{12-Byte}&
\textbf{4-Byte}\\
\hline
Unknown&
Block count\\
\hline
\end{tabular}
\begin{tabular}{|>{\centering\arraybackslash}p{240pt}|}
\textbf{32-Byte}\\
\hline
SHA256 payload\\
\hline
\end{tabular}
\begin{tabular}{|>{\centering\arraybackslash}p{57pt}|>{\centering\arraybackslash}p{57pt}|>{\centering\arraybackslash}p{54pt}|>{\centering\arraybackslash}p{54pt}|}
\textbf{4-Byte}& 
\textbf{4-Byte}&
\textbf{4-Byte}&
\textbf{4-Byte}\\
\hline
Name&
Start offset&
Output size&
Attributes\\
\hline
\end{tabular}
\label{tab1}
\end{table}

DJI products use proprietary encryption and signatures to ensure confidentiality and integrity[24]. So, according to the above Table 3, the part of the DJI drone firmware file header that we should be mainly interested in is the "Encryption key Identifier" part with a size of 4-Byte. An identifier is written in this area to identify which key each firmware file was encrypted with. The decryption key is different depending on the identifier written in the area. In addition, DJI's firmware performs verification through the above auth key identifier, and only after the verification process is it executed after decryption.

In order to obtain the decryption key corresponding to the identifier code, we collected and analyzed previous projects and papers related to DJI drone firmware decryption. Our analysis revealed a variety of related research projects, but we focused our attention on the dji-firmware-tools project by a specific hacking group[23]. In this project, the decryption key was obtained through side-channel analysis to analyze the firmware of DJI drones, and it was found to be successful in decrypting a significant number of DJI drone firmware[22], [23]. We decrypted the DJI drone firmware according to the "Encryption key Identifier" collected from the publicly available information of the project. First, as a sample, we manually decrypted a partial firmware file instead of the entire firmware file. As a result of the decryption, we succeeded in decrypting most of the sample files. The following Table 4 shows the key identifiers we used for decryption.\\

\begin{table}
\caption{Firmware decryption identifiers and keys}
\label{table}
\setlength{\tabcolsep}{3pt}
\begin{tabular}{|>{\centering\arraybackslash}p{57pt}|>{\centering\arraybackslash}p{57pt}|>{\centering\arraybackslash}p{114pt}|}
\hline
\textbf{Key Identifier}& 
\textbf{Key Version}&
\textbf{Key Value}\\
\hline
PRAK& 
PRAK-2017-01&
40000000c3151641157d3…\\ 
PRAK&
PRAK-2017-08&
40000000dbe15b5badcde…\\ 
PRAK&
PRAK-2017-12&
40000000c3151641157d3…\\
PRAK&
PRAK-2018-01&
400000008f73897091b44…\\
PRAK&
PRAK-2019-09&
40000000a1f987bf9fd539…\\ 
PRAK&
PRAK-2020-01&
40000000c73fb7ba092e1f…\\
\hline
 \multicolumn{3}{|c|}{...}\\ 
\hline
 RREK& RREK-2017-01&0x37, 0xD6, 0xD9, …
\\ 
 IAEK& RIEK-2017-01&0xF1, 0x69, 0xC0, …
\\ 
 IAEK& IAEK&0x89, 0x9D, 0x1B, …\\ \hline
\end{tabular}
\label{tab1}
\end{table}

We manually decrypted the firmware. The process of manually decrypting the firmware requires entering the decryption keys corresponding to the "Encryption key Identifier" in the firmware file header one by one and performing repetitive operations until the decryption is successful. However, since the "Encryption key Identifier" varies for each product, version, etc. it is inefficient to manually decrypt one by one. Therefore, we developed an automation module for decrypting each firmware to build an automatic firmware analysis system.

First, we constructed a dictionary that maps "Encryption key Identifier" and the corresponding decryption key for each identifier. The mapping resulted in a dictionary with a 1-to-n structure, where one "Encryption key Identifier" is mapped to one or more decryption keys. Based on the configured dictionary, the firmware decryption module operates in the following sequence:\\

1) Parses the "Encryption key Identifier" described in a specific area of the entire firmware file.

2) Extract the decryption key from the dictionary based on the parsed identifier.

3) Attempt decryption using the extracted decryption key (or list of keys).\\

We attempted to decrypt all firmware files with the firmware decryption module we developed to work in the order listed above. As a result, we succeeded in decrypting 4,233 files (91.48\%) out of a total of 4,627 encrypted firmware files. The following Figure 4 shows an example of our successful decryption.

\Figure[t!](topskip=0pt, botskip=0pt, midskip=0pt){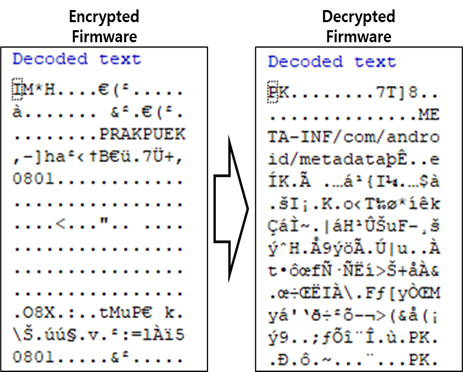}
{Decrypted DJI drone firmware\label{fig4}}

In addition, the files that failed to decrypt were mostly configuration files in .cfg format and initialization files in .ini format with only the signature value in the binary header and no encryption, as shown in Figure 5. There was no correlation to a specific product or version.

\Figure[t!](topskip=0pt, botskip=0pt, midskip=0pt)[scale=0.52]{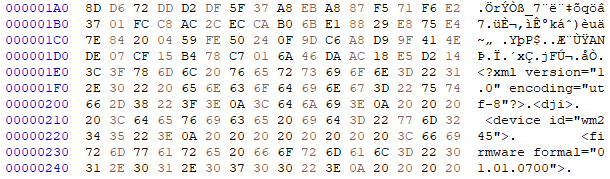}
{Unencrypted configure files\label{fig5}}

\subsubsection{Dynamic analysis using the dorne}
This section describes how to perform dynamic analysis on drone devices. Since drones are embedded systems, no new techniques are required to analyze drone firmware, and existing dynamic analysis methods can be used. In other words, drones can be analyzed in the same way that the firmware of IoT devices, one of the subcategories of embedded systems, is analyzed. To perform dynamic analysis of drone firmware, existing publicly available dynamic analysis tools can be used by embedding them in an emulator or directly in the drone system. 

We first performed dynamic analysis of drone firmware by modifying an existing dynamic analysis tool to run on a drone system. We performed a cross-compilation of existing analysis tools to prepare them to run on the drone system. Below are the commands we used to cross-compile the dynamic analysis tool.\\

\textbf{\$ arm-linux-gnueabi-gcc -march=armv5te -mfloat-abi=soft -I/usr/arm-linux-gnueabi/include -static afl-fuzz.c afl-fuzz}\\

After cross-compilation, we were able to insert the cross-compiled analysis tools into the drone system via ADB and run them against the internal binaries. We were also able to connect to the analysis computer and perform debugging using the GDB server that exists inside the drone. Figures 6 and 7 show GDB server and fuzzing performed using a DJI drone.

\Figure[t!](topskip=0pt, botskip=0pt, midskip=0pt)[scale=0.45]{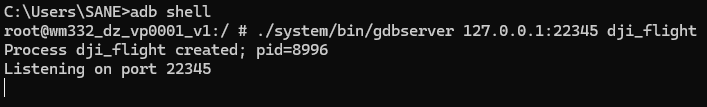}
{gdbserver on a DJI drone using ADB shell\label{fig6}}

\Figure[t!](topskip=0pt, botskip=0pt, midskip=0pt)[scale=0.43]{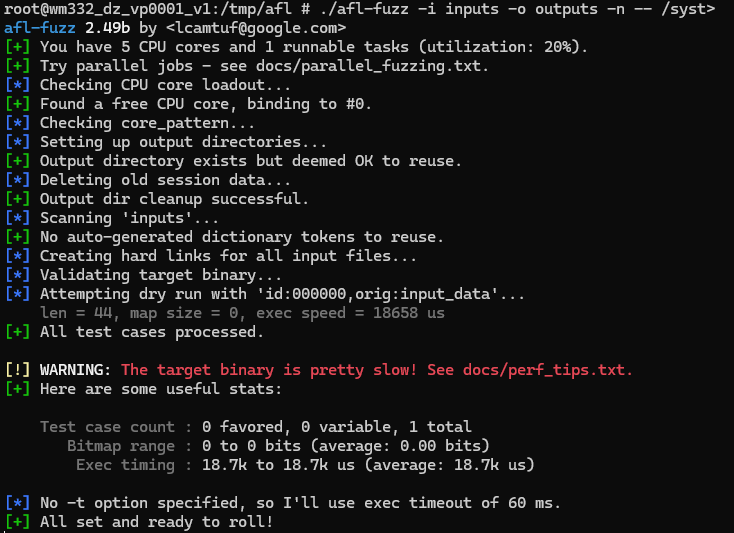}
{Fuzzing DJI drone firmware using drone{(DJI Phantom 4 Pro)}\label{fig7}}

However, there were several problems with running existing analysis tools directly on the drone system. While a sufficient amount of time is required to analyze a program's vulnerabilities, drones cannot be plugged directly into a power line for electricity, and even a fully charged battery can only operate for a limited amount of time due to capacity limitations. These limitations on system uptime make it difficult to install and run dynamic analysis tools directly on the drone. Even if the drone system has a good power supply, the limited hardware environment, including the limited memory capacity and processing power of the drone body, prevents the efficient execution of dynamic analysis tools, which affects the effectiveness of the analysis. For these reasons, it takes a long time to perform dynamic analysis by installing dynamic analysis tools directly on the drone system, or there are limitations in the dynamic analysis of large files. In addition, it is difficult to use existing analysis methods and tools due to the unique characteristics of drones, such as various systems such as motors, cameras, etc. that are connected and operated, and processes that operate in real time. Therefore, it is necessary to build an environment that is used by the drone system or to perform the analysis in the drone system's operating environment itself.

In the case of the drone system operating environment itself, it has a very harsh environment. The DJI Phantom 4, the main target of this study, does not have detailed hardware specifications, but the Pixhawk V6X Autopilot, Pixhawk's latest flight controller, which is commonly used for DIY (Do It Yourself) drone production, has a 480Mhz processor and 1MB of memory. This drone system environment is not suitable for dynamic analysis as it requires a lot of system resources. Therefore, we prepared an embedded board with a similar environment to the drone system, but with much higher performance. We used the Beaglebone Black embedded board to configure the analysis environment, and it shows a high performance of about 1090 times (about 2.13 times processor and 512 times memory) compared to the Pixhawk V6X Autopilot above, considering the main resources required for dynamic analysis: processor and memory. In this way, we have solved the problem of harsh environments by using embedded environments that can replace drone systems and have higher performance.

\Figure[t!](topskip=0pt, botskip=0pt, midskip=0pt)[scale=0.6]{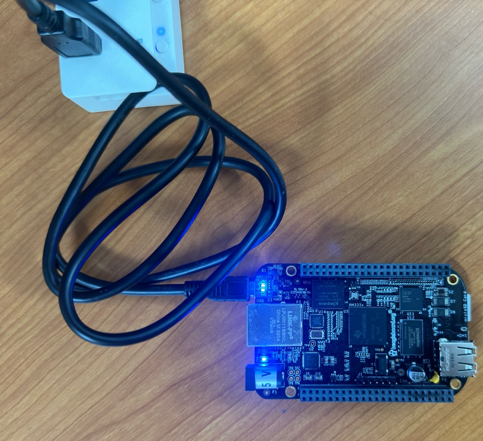}
{Embedded Board {(Beaglebone Black)}\label{fig8}}

\subsubsection{Installing dependent libraries}
This section describes building an emulation environment for dynamic analysis of decrypted firmware. Firmware analysis involves the use of tools such as emulators and debuggers, and the environment must be configured so that the commands in the analysis tool work properly. Since firmware is highly hardware dependent, the same operating environment as that used by the system to be analyzed must be configured to run the analysis tool. The DJI drone system uses the Android operating system, and the firmware binary does not have PIE (Position Independent Executables), which is mandatory for security reasons in Android 5.0 (Lollipop) and later. Therefore, we need to use an Android version lower than 5.0 to build an emulation environment for DJI drone firmware binaries.

In addition, since the DJI drone system uses a customized Android environment, it requires libraries that are used independently. Therefore, we need to check the files used in addition to the libraries used in the Android environment and identify the dependency relationships for each file to the required libraries. We used the command "ldd" command to identify the required libraries and dependencies, and checked the required files for each binary, as shown in Figure 9. We then, we accessed the internal system of the DJI drone using ADB, extracted the required library files, and installed them on the embedded board to complete the environment setup for dynamic analysis.

\Figure[t!](topskip=5pt, botskip=5pt, midskip=5pt)[scale=0.42]{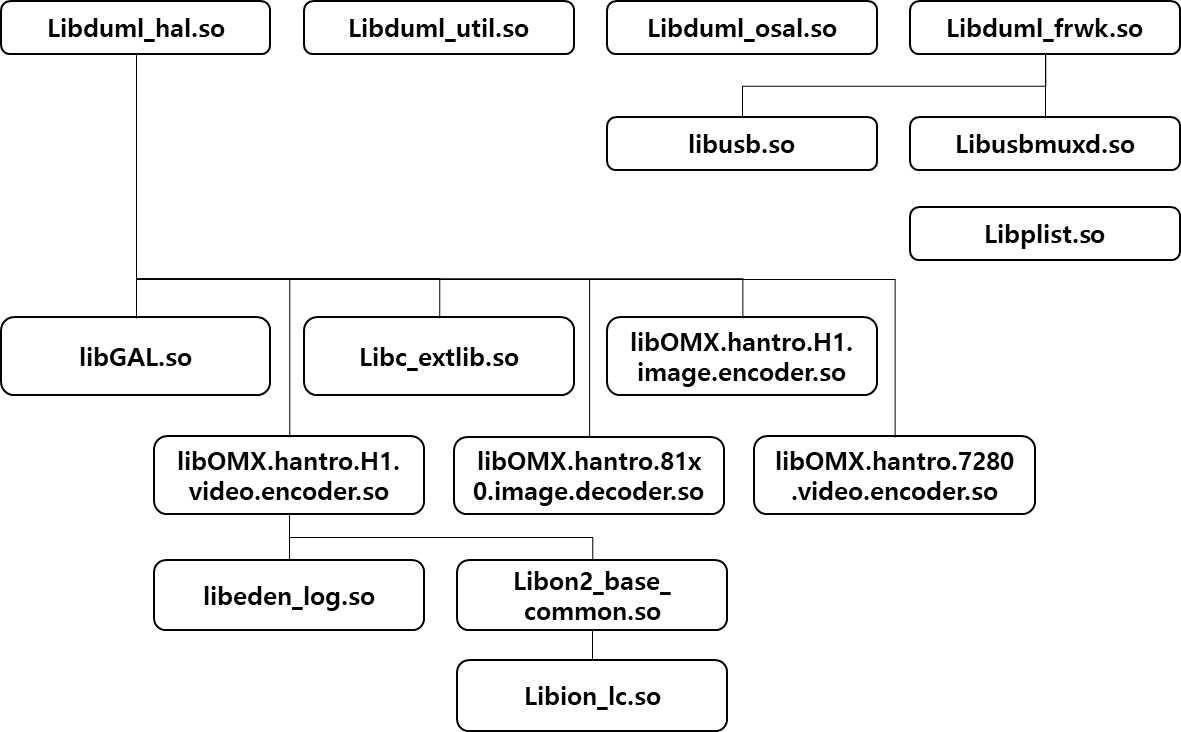}
{Dependencies of libraries used by DJI drones environment\label{fig9}}

\subsubsection{Installing fuzzer}
To undertake an automated analysis of DJI firmware, it is imperative to employ tools like AEG and fuzzer. American Fuzzy Lop (AFL), widely adopted for such purposes, compiles source code via afl-gcc for fuzzing, or resorts to black box fuzzing through user space emulation for binary code. However, given the DJI firmware's source code is not open to the public and thus unobtainable, analysis is confined to the firmware's binary form.

Upon examining the decrypted DJI firmware, it became apparent that the majority of the binaries are ARM architecture-based. This necessitates the use of tools compatible with various architectures, including ARM, Aarch64, and PowerPC. Considering the focus on firmware, vulnerability analysis tools that necessitate source code, such as honggfuzz[27], KLEE[28], and LibFuzzer[29], are not applicable. Consequently, suitable tools for this research include AFL, Radamsa[30], and angr[31], all of which support binaries across different platforms and architectures. Radamsa, however, falls short for this project since it modifies the structure of existing binaries and relies on manual post-execution error checks and analyses. Angr also presents challenges as it requires setting up a simulation manager for each binary to determine probable vulnerability targets, rendering it less viable for extensive automated analysis.

AFL was chosen, a cross-platform fuzzer, alongside an AFL-equipped embedded board, to conduct the firmware's automated analysis. AFL's shared memory capabilities facilitate code coverage tracking, performance optimization, and novel input identification. However, the shared memory API it typically uses is restricted to the Android operating systems running drone systems, requiring an alternative API. Because AFL is primarily designed for Linux or Unix-based systems, it is incompatible with Android-based embedded boards. AFL's reliance on system-specific libraries and debugging tools means that the embedded AFL binary cannot run on different operating systems without modifications that take into account the unique features and constraints of the Android platform. To overcome these shared memory API challenges, an Android-specific AFL version, Android-AFL[32], was discovered and subsequently deployed within the Beaglebone Black embedded board environment. Tailored to the board's ARM Cortex-A8 processor, a cross-compilation environment was built and configured for this precise architecture. Thus, we adapted Android-AFL for use on embedded boards and built the platform to perform dynamic analysis in a separate analysis environment, as shown in Figure 11.
 
\Figure[t!](topskip=0pt, botskip=0pt, midskip=0pt)[scale=0.29]{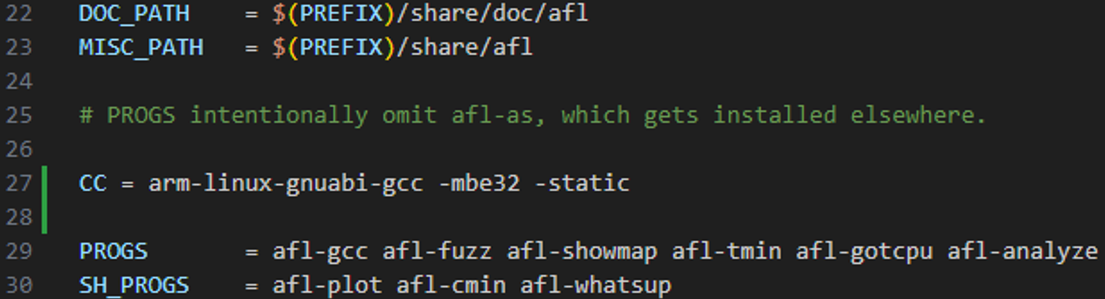}
{Modifying Makefile for cross-compiling\label{fig10}}

\Figure[t!](topskip=0pt, botskip=0pt, midskip=0pt)[scale=0.53]{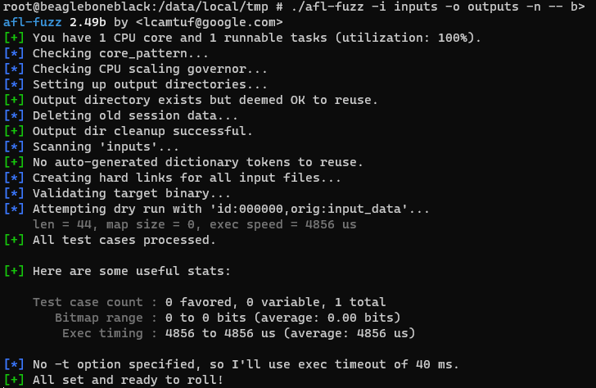}
{Fuzzing DJI drone firmware using embedded boards\label{fig11}}

%수정 "DTA: Run TrustZone TAs Outside the Secure World for Security Testing" 참고

\section{Evaluation}
In this section, we present a comprehensive evaluation of our built firmware analysis platform on DJI drones to illustrate the merits of our findings. First, we evaluate the differences between current firmware analysis tools and our built DJI drone firmware dynamic analysis platform. Second, we conduct experiments on the speed difference between our analysis and the built platform using drone devices with Android-AFL. In summary, we aim to address the following research questions:\\

 \textbf{RQ1 : }How effective is the proposed dynamic analysis methodology in identifying security vulnerabilities in drone firmware? (Section IV-B)
 
 \textbf{RQ2 : }What differentiated advantages does the construction of a dynamic analysis platform for DJI drone firmware offer compared to current firmware analysis tools? (Section IV-C)
 
\subsection{Environmental setup}
The experiments were conducted on an embedded board called the Beaglebone Black, on which the DJI drone firmware analysis platform was built. To run the DJI firmware within the platform, the internal operating system used Android version 4.4, Kitkat. We added libraries that are proprietary to DJI drones to that environment, and modified some code to use Android-AFL in that environment. Table 5 describes the configured environment.

\begin{table}
    \centering
\caption{Experimental setup}
\label{tab:my_label}
    \begin{tabular}{|>{\centering\arraybackslash}p{45pt}|>{\centering\arraybackslash}p{74pt}|>{\centering\arraybackslash}p{84pt}|} 
    \hline 
         \textbf{Classification}&  \textbf{Configuration}& \textbf{Parameters}\\
         \hline
         &  CPU& Cortex A8\\ 
            Hardware& Memory&512MB\\ 
            & Flash&2GB\\
        \hline
         &  Operating system& Android 4.4.4 (Kitkat)\\ 
         Software&  Kernel version& 3.8.13+\\
        & Processor&ARM\\
  \hline
    \end{tabular}
\end{table}

\subsection{Evaluate the performance of your deployed firmware analysis platform (RQ1)}
We were able to dynamically analyze the binary of the DJI drone system within the Beaglebone Black embedded board environment without any problems. Fuzzing exercises performed directly on the drone body and through our custom embedded board are shown in Figures 7 and 10, respectively. Our results show that while the drone body took 18.7k microseconds to execute and complete a test case, the embedded board took only 4,856 microseconds for the same file, a remarkable speed improvement of approximately 3.85 times per test case. This efficiency gain is expected to be even more significant for processes with large computation cycles.

\begin{figure}
    \centering
    \includegraphics[width=1\linewidth]{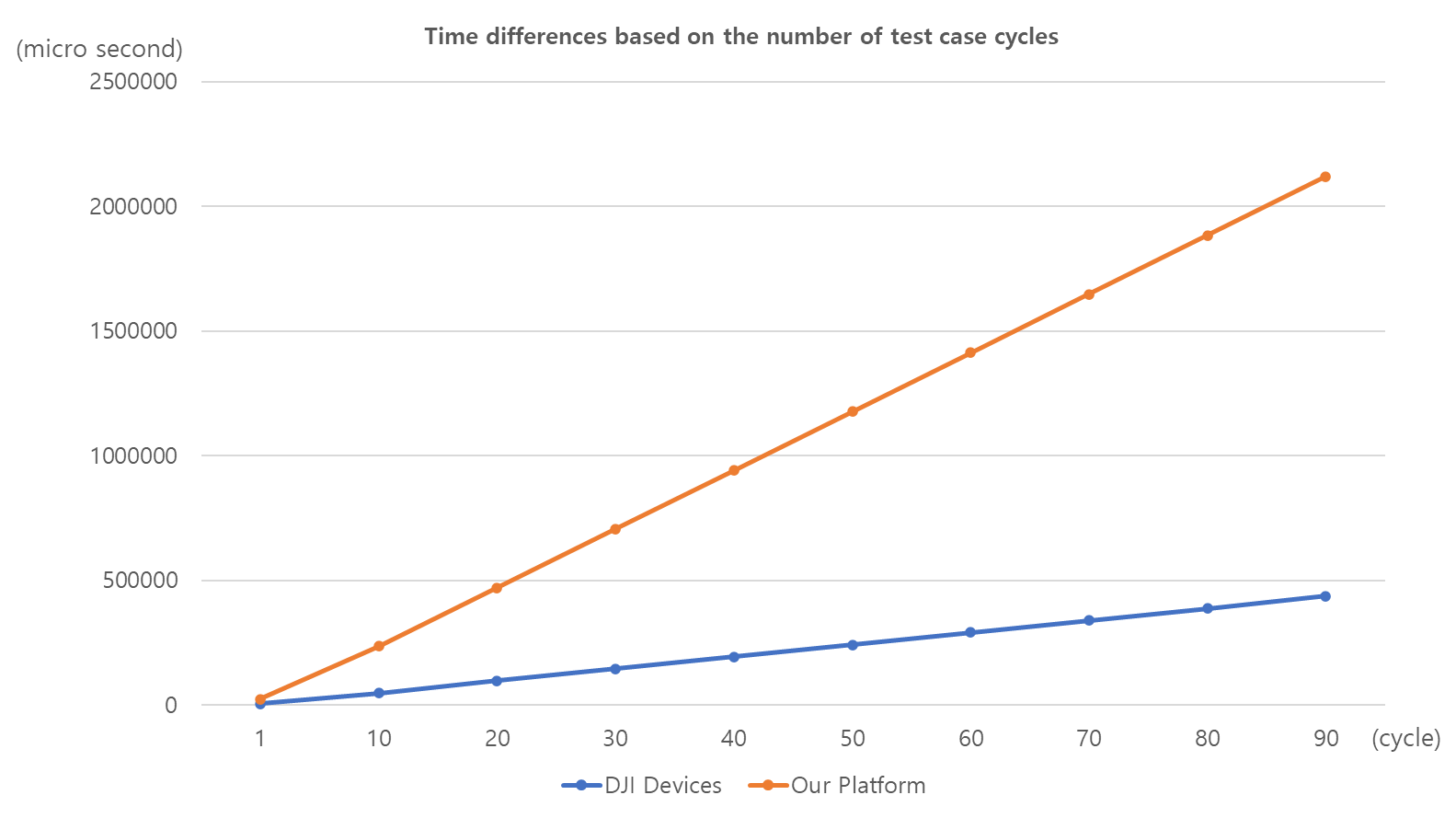}
    \caption{Time differences based on the number of test case cycles}
    \label{fig:enter-label}
\end{figure}

\subsection{Differences between existing research and our proposed method (RQ2)}
This study proposes the current gaps in dynamic drone firmware analysis by proposing a drone firmware analysis methodology. The proposed methodology includes a firmware analysis solution that starts with an automated decryption process. This innovative process eliminates the need for manual decryption, which is not only time-consuming but also prone to errors, thereby increasing the reliability and speed of the initial stages of firmware analysis.

We also present the development of an analysis environment that closely mimics the operating conditions of actual drone systems. Using embedded boards, our method simulates the hardware of the drone, allowing for a more accurate and nuanced analysis of firmware behavior in real-world scenarios. This environment provides a solid foundation for the dynamic analysis tools to operate on, ensuring that the analysis is as close as possible to the actual operating conditions of the drone.

A cornerstone of our proposed methodology is the application of Android-AFL for dynamic analysis. This application represents a significant leap forward in the automated detection of vulnerabilities in drone firmware. Android-AFL allows us to perform exhaustive fuzzing operations that not only detect vulnerabilities, but also assess the potential impact of such vulnerabilities in operational environments. This is a significant advance over traditional methods, which typically focus on static analysis and often miss the subtleties of vulnerabilities that could be exploited in the field.

Our approach significantly streamlines the vulnerability assessment process by incorporating AFL fuzzing for automated dynamic analysis. The AFL tool brings a high level of sophistication to our analysis by automating the fuzzing process and ensuring comprehensive coverage of the firmware code. This automation dramatically reduces the labor-intensive aspects of dynamic analysis, allowing researchers to spend more time interpreting and mitigating discovered vulnerabilities.

Finally, the proposed method achieves efficiency. Where traditional methods rely heavily on manual processes, our automated dynamic analysis framework offers a significant reduction in time and effort, bringing drone security research into a new era of efficiency and effectiveness. By reducing manual effort, we unlock the potential to scale analysis across multiple firmware versions and drone models, which is critical to staying ahead of the evolving threat landscape.

Table 6 succinctly captures the transformative effect of our proposed method over traditional approaches. It highlights the systematic and strategic improvements our methodology brings to each phase of drone firmware analysis, setting a new standard in the field.

\begin{table}
    \centering
\caption{Traditional and proposed method}
\label{tab:table}
    \begin{tabular}{|>{\centering\arraybackslash}p{35pt}|>{\centering\arraybackslash}p{84pt}|>{\centering\arraybackslash}p{84pt}|} \hline 
         \textbf{Aspect}&  
         \textbf{Traditional Methods}& 
         \textbf{Proposed Method}\\ \hline 
         Approach&  
         Mainly reliant on manual analysis with limited tool usage& 
         Development of an automated dynamic analysis methodology using embedded boards and tools\\ \hline 
         Firmware Acquisition and Decryption& 
         Manual collection and decryption of firmware, time-consuming and prone to errors& 
         Development of an automated module for efficient firmware collection and decryption\\ \hline 
         Analysis Environment Setup&  
         Dependent on limited hardware resources and environment, difficulties in managing necessary libraries for firmware execution	
         & Use of embedded boards to create an analysis environment similar to actual drone systems, solving library dependencies\\ \hline 
         Vulnerability Analysis Method&  
         Relies primarily on static and protocol analysis, potentially missing vulnerabilities in real-world production environments& 
         Use of AFL fuzzing for automated dynamic analysis, identifying vulnerabilities in operational environments\\ \hline 
         Automation and Efficiency&  
         Limited automation of the analysis process, potentially time-consuming and labor-intensive	& 
         High degree of process automation leading to significant time and effort reduction, achieving high efficiency\\ \hline
    \end{tabular}

\end{table}

\section{Conclusion and Future Work}
We are by addressing the challenges associated with dynamic analysis of drone firmware, particularly that of DJI drones, we have introduced a comprehensive methodology that not only streamlines firmware decryption, but also facilitates automated vulnerability analysis. The creation of a dedicated embedded board, along with the development of an automated decryption module, represents a significant advancement in the field. This setup enables the efficient use of automated analysis tools, overcoming the limitations of traditional methods, which often rely on manual processes and are time-consuming.

Our approach improves the efficiency of vulnerability analysis in drone firmware by automating the decryption process for encrypted firmware files and building a drone simulation environment for dynamic analysis will contribute to the advancement of drone security research. Given the increasing reliance on drones in various fields and their potential security vulnerabilities, we provide an automated vulnerability analysis methodology that allows for a more thorough and rapid security assessment of drone firmware. In addition, the methodology and tools developed through this research provide a replicable model for future research that can be extended not only to DJI drones, but also to other UAV systems using the Android environment. This versatility increases the broader applicability of our findings and contributes to the development of safer drone technologies in general. Based on these contributions, it is clear that our work is a critical step towards solving the complex problem of drone firmware security. The automated process introduced here significantly reduces the barriers to comprehensive security analysis, enabling more effective identification and mitigation of vulnerabilities in drone systems.

%reference 작성

\begin{IEEEbiography}[{\includegraphics[width=1in,height=1.25in,clip,keepaspectratio]{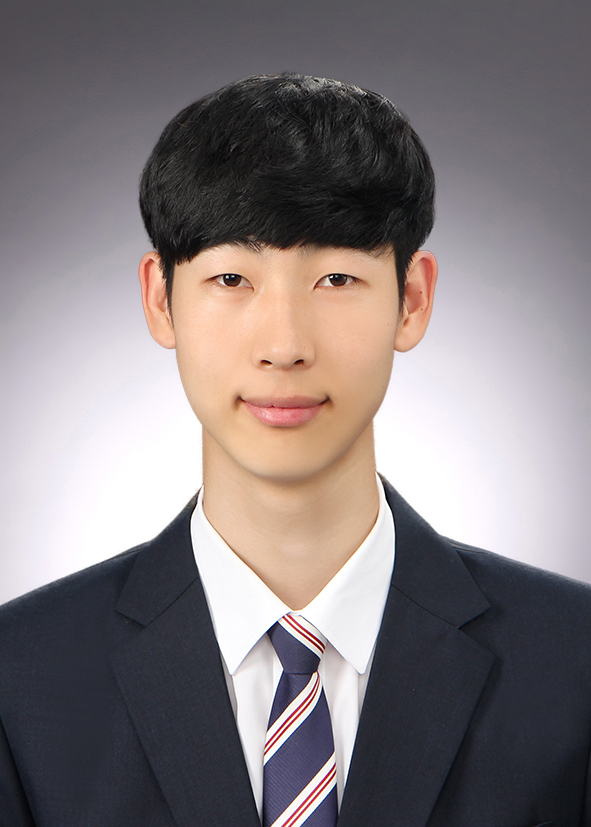}}]{Yejun Kim} (Member, IEEE) received the B.S.degrees in information security from Soonchunhyang University, South Korea, in 2019. He is a 2019 graduate of the Integrated Master's and Doctoral Program in Information Security at the Graduate School of Information Security, Korea University, Seoul, Korea. 

His current research interests include security engineering, SDLC, security testing, and security assessment.
\end{IEEEbiography}

\begin{IEEEbiography}[{\includegraphics[width=1in,height=1.25in,clip,keepaspectratio]{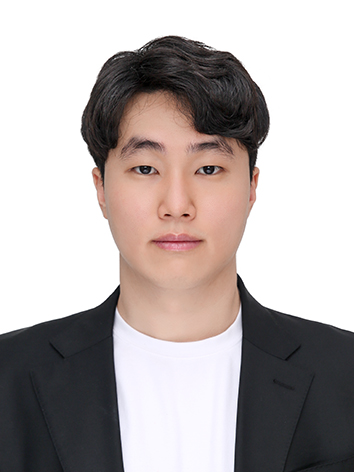}}]{Kwangsoo Cho} (Member, IEEE) was born in , Daejun, Chungcheongnam-do, Republic of Korea in 
1996. He received the B.S. in computer engineering from the Hoseo University, Asan-si, Chungcheongnam-do, in 2019 and the M.S. degree in school of information security from the Korea University, Seoul, in 2021.

His research interest includes Security Software Development Lifecycle, Risk Management, and Threat Modeling.
\end{IEEEbiography}

\begin{IEEEbiography}[{\includegraphics[width=1in,height=1.25in,clip,keepaspectratio]{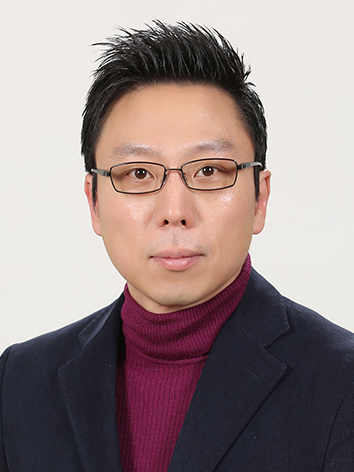}}]{Seungjoo Kim} (Member, IEEE) received the B.S., M.S., and Ph.D. degrees in information engineering from Sungkyunkwan University, South Korea, in 1994, 1996, and 1999, respectively. Since 2011, he has been a Professor with the School of Cybersecurity, Korea University. For the past 7 years, he was an Associate Professor at Sungkyunkwan University and has 5 years of back ground of the Team Leader of Korea Internet and Security Agency (KISA). In addition to being a professor, he is a vice president for Digital Information of Korea University from 2023, a dean of School of Smart Mobility, Korea University from 2022, a head of SANE(Security Assessment aNd Engineering) Lab, an adviser of undergraduate hacking club  'CyKor (DEFCON CTF 2015 \& 2018 winner)'  at the School of Cybersecurity, Korea University from 2011 to February 2020. He solicits research on a broad range of topics relating to secure systems development such as DevSecOps, security assessment such as Common Criteria, CMVP, SSE-CMM, RMF A\&A etc., and blockchain.

\end{IEEEbiography}

\EOD

\end{document}